# Enhanced Sentiment Analysis of Iranian Restaurant Reviews Utilizing Sentiment Intensity Analyzer & Fuzzy Logic

Shayan Rokhva, Babak Teimourpour*, Romina Babaei

Department of Information Technology Engineering, Faculty of Industrial and Systems Engineering, Tarbiat Modares University, Tehran, Iran [affiliation for all authors]

Shayanrokhva1999@gmail.com & Shayan_rokhva@modares.ac.ir [ORCID ID: 0009-0006-7558-0312]

Babaktei@gmail.com & B.teimourpour@modares.ac.ir [ORCID ID: 0000-0002-9286-2286]

Rominababaeii79@gmail.com [ORCID ID: 0009-0003-3620-8903]

## ABSTRACT:

This research presents an advanced sentiment analysis framework studied on Iranian restaurant reviews, combining fuzzy logic with conventional sentiment analysis techniques to assess both sentiment polarity and intensity. A dataset of 1266 reviews, alongside corresponding star ratings, was compiled and preprocessed for analysis. Initial sentiment analysis was conducted using the Sentiment Intensity Analyzer (VADER), a rule-based tool that assigns sentiment scores across positive, negative, and neutral categories. However, a noticeable bias toward neutrality often led to an inaccurate representation of sentiment intensity. To mitigate this issue, based on a fuzzy perspective, two refinement techniques were introduced, applying square-root and fourth-root transformations to amplify positive and negative sentiment scores while maintaining neutrality. This led to three distinct methodologies: Approach 1, utilizing unaltered VADER scores; Approach 2, modifying sentiment values using the square root; and Approach 3, applying the fourth root for further refinement.  A Fuzzy Inference System incorporating comprehensive fuzzy rules was then developed to process these refined scores and generate a single, continuous sentiment value for each review based on each approach. Comparative analysis, including human supervision and alignment with customer star ratings, revealed that the refined approaches significantly improved sentiment analysis by reducing neutrality bias and better capturing sentiment intensity.  Despite these advancements, minor over-amplification and persistent neutrality in domain-specific cases were identified, leading us to propose several future studies to tackle these occasional barriers. The study's methodology and outcomes offer valuable insights for businesses seeking a more precise understanding of consumer sentiment, enhancing sentiment analysis across various industries.

## KEYWORDS:

Sentiment Analysis, Fuzzy Logic, Restaurant Reviews, Sentiment Intensity Analyzer, Fuzzy Inference System (FIS), Natural Language Processing (NLP)

## 1. Introduction

### 1.1. The concept and importance of study

Sentiment analysis typically refers to the examination of opinions, emotions, and attitudes expressed in text, aiming to classify them as positive, negative, or neutral (Haque, 2014; Vashishtha *et al.*, 2023). In today's fiercely competitive, data-centric world, even small firms leverage this technique to understand customer attitudes, offer better recommendations, and enhance customer experience, ultimately maximizing profits and expanding their market share (Ahamed *et al.*, 2019; Rokhva *et al.*, 2024; Rokhva and Teimourpour, 2025). For instance, social media platforms use sentiment analysis to gauge public opinion on issues, brand perceptions, and emerging trends, which helps remove inappropriate content and enables rapid responses to viral posts (Ahamed *et al.*, 2019). Similarly, online retailers benefit from detecting recurring complaints and praise, helping to refine product suggestions, optimizing inventory management, and improving customer service (Cirqueira *et al.*, 2020; Souza *et al.*, 2015). In the restaurant industry, where customer reviews are pivotal, sentiment analysis helps understand dining experiences and guide improvements in menus, ambiance, and overall service. It also enables food recommender systems to offer refined suggestions, enhancing satisfaction, overall experience, and customer loyalty (Reddy and Priyadarsini, 2025; Rokhva *et al.*, n.d.)

However, the real challenge of analyzing sentiment lies in effectively managing human feedback's inherent uncertainty, ambiguity, and complexity. Opinions can be multifaceted, reflecting complete satisfaction, partial contentment, or even a mixture of satisfaction and dissatisfaction. Whether we use deep learning or conventional NLP and sentiment approaches, this process cannot be done by simply assigning a single label to each record of data without considering the intensity and complexity inherent in this context (Ahamed *et al.*, 2019; Vashishtha *et al.*, 2023). In response, fuzzy logic provides an elegant solution by replacing rigid binary categorizations (0 or 1) with a continuum of values between 0 and 1 that reflect varying degrees of truth. This method is particularly advantageous in sentiment analysis, where texts seldom convey purely positive or purely negative sentiments (Matiko *et al.*, 2014; Vashishtha *et al.*, 2023). For example, a restaurant review might commend the exceptional service while simultaneously criticizing the food quality. Fuzzy logic adeptly captures these nuances by assigning values that reflect different levels of satisfaction and discontent through systematic computation and logical assessment (Cardone *et al.*, 2023; Yan *et al.*, 2022).

Building upon fuzzy logic, fuzzy inference systems (FIS) provide a robust framework for decision-making amid uncertainty. These systems integrate multiple criteria to emulate human expert judgment, thereby facilitating and speeding up decision-making with a non-binary perspective (Abbasi *et al.*, 2025; Dhyani *et al.*, 2022). In the realm of sentiment analysis, FIS can consolidate various sentiment indicators extracted from text and synthesize them into a single, continuous score ranging from 0 to 1. As previously cited, this score not only represents the overall sentiment but also quantifies its intensity. This automated, expert-like evaluation—considering dimensions such as intensity, polarity, and context—is particularly valuable in domains like social media, online retail, and restaurants, where actionable insights enable decision-makers to fine-tune their strategies (Jiang *et al.*, 2024; Zhu and Ding, 2025).

In summary, the fusion of conventional methods of sentiment analysis and fuzzy logic offers enhanced and more accurate sentiment analysis. Conventional methods can interpret the text and partially extract good features from data, and fuzzy logic and fuzzy inference systems can combine these features to handle uncertainties and complexities inherent in human emotions (Li *et al.*, 2024; Vashishtha *et al.*, 2023). This integrated approach significantly boosts the effectiveness of sentiment analysis, rendering it an indispensable tool for enhancing customer satisfaction, optimizing service delivery, and driving success in competitive sectors.

### 1.2. Brief survey of similar works

Over the years, researchers have explored different computational techniques to improve sentiment classification accuracy, with a growing interest in fuzzy logic due to its ability to handle the inherent ambiguity of natural language (Vashishtha *et al.*, 2023). Early approaches to sentiment analysis relied heavily on lexicon-based. Then, with the hype around AI, machine learning perspective integrated into numerous applications from computer vision, natural language processing, graph-based applications, and sentiment analysis(Shamami, Farahzadi, *et al.*, 2024; Shamami,

Teimourpour, *et al.*, 2024). Despite their robustness, their applications in sentiment analysis struggled with uncertain and imprecise sentiment expressions, mostly failing to determine intensity (Vashishtha *et al.*, 2023).

To address these challenges, fuzzy logic has been increasingly integrated into sentiment analysis models. A comprehensive review in this area categorizes fuzzy-based approaches into five primary classes: sentiment cognition from words and phrases, fuzzy-rule-based methods, neuro-fuzzy networks, and fuzzy emotion recognition (Haque, 2014). This classification highlights how fuzzy logic enhances sentiment analysis by addressing vagueness and subjectivity in textual data.

Several studies have attempted to integrate fuzzy logic with deep learning and neural networks to improve sentiment classification performance. For instance, the integration of fuzzy logic into graph convolutional networks (GCNs) has been proposed to reduce sentiment ambiguity in the level of sentences. The Fuzzy Graph Convolutional Network (FGCN) method enhances knowledge representation under uncertainty, showing superior accuracy over traditional GCNs (Phan and Nguyen, 2024). Similarly, fuzzy-deep neural networks that incorporate multimodal data, such as text, audio, and images, have been explored to enhance sentiment classification accuracy. These models integrate a dual-attention mechanism that refines feature extraction, proving effective in capturing the intricacies of human emotions (Wang *et al.*, 2025).

Aspect-based sentiment analysis (ABSA) has also benefited from fuzzy logic, particularly in e-commerce and customer review analysis. Traditional machine learning approaches in ABSA often fail to accurately capture nuanced sentiments related to specific product attributes. A hybrid model combining LSTM networks with fuzzy logic has been proposed to classify sentiment into multiple categories, achieving higher accuracy compared to traditional methods (Zhu and Ding, 2025). Likewise, a fuzzy recommendation system has been developed to dynamically predict customer interests by integrating sentiment analysis with ontology alignment, ensuring more accurate product recommendations based on evolving user preferences (Karthik and Ganapathy, 2021).

In decision-making contexts, fuzzy logic has proven beneficial in ranking products, services, and suppliers based on sentiment analysis. For instance, supplier selection models that integrate aspect-level sentiment analysis with fuzzy multi-attribute decision-making (MADM) have demonstrated improved accuracy in evaluating suppliers by incorporating subjective user opinions alongside objective criteria (Sun *et al.*, 2024). Similarly, intuitionistic and hesitant fuzzy methods have been employed to rank tourist attractions and restaurants based on online reviews, highlighting the flexibility of fuzzy logic in multi-criteria decision-making scenarios (Qin *et al.*, 2022).

Moreover, sentiment analysis has been leveraged in specialized applications such as fake news detection and opinion mining in social media. The integration of fuzzy logic with deep learning models like Bidirectional Long Short-Term Memory (BiLSTM) has shown promising results in analyzing student feedback on e-learning and detecting misinformation during the COVID-19 pandemic. In particular, fuzzy-enhanced models have outperformed traditional sentiment classifiers by effectively handling ambiguous and informal text (Alzaid and Fkih, 2023).

Another emerging trend is the use of sentiment analysis for evaluating new energy vehicles (NEVs) and consumer preferences. By employing fuzzy logic to process unstructured online reviews, researchers have developed picture fuzzy entropy-based weighting methods to quantify consumer sentiment towards NEVs. This approach has led to more robust evaluations that account for consumer psychology and bounded rationality in decision-making (He and Wang, 2023). Additionally, scalable ensemble methods employing fuzzy inference mechanisms have demonstrated improvements in processing large-scale social media data for sentiment estimation (Isikdemir and Yavuz, 2022).

Overall, integrating fuzzy logic in sentiment analysis has significantly enhanced the ability to process and interpret subjective opinions. By combining fuzzy logic with deep learning, neural networks, and multi-criteria decision-making frameworks, researchers have developed more sophisticated models capable of handling ambiguity, improving classification accuracy, and supporting decision-making in various domains. Future research should focus on further refining these approaches, particularly in multimodal sentiment analysis and real-time business intelligence and social media monitoring applications.

### 1.3. Research efforts and contributions

The literature review revealed a gap in the sentiment analysis of Persian restaurants, particularly when considering the varying intensity and complexity of sentiments. To address this gap, we developed a local dataset comprising reviews from multiple Iranian restaurants, which includes Persian comments, corresponding accurate AI translation, the number of items purchased, and star ratings reflecting customer satisfaction. We then conducted an in-depth study of sentiment analysis for these restaurants by integrating a Sentiment Intensity Analyzer and fuzzy logic principles and techniques.

The current study makes the following key contributions:

I. **Creation of a Local Dataset**

   We compiled a unique dataset focused on Iranian restaurants, incorporating Persian reviews, accurate English translation reviews by AI, purchase quantities, and star ratings as proxies for customer satisfaction.

II. **Application of Sentiment Intensity Analyzer:**

   We employed the Sentiment Intensity Analyzer (a rule-based VADER tool from NLTK) to assign three sentiment scores (positive, negative, and neutral) to each review. This set of scores, which sum to one, provides an initial, rough assessment of sentiment for each record, later refined and processed in our study.

III. **Refinement of Sentiment Scores Using Fuzzy Logic Principles:**

   Recognizing that the initial scores were often biased towards neutrality, we introduced two refinement approaches. In the first approach, we amplified the positive and negative scores by taking their square roots while leaving the neutral score unchanged. In the second approach, we applied the fourth root to the positive and negative scores, again keeping the neutral score untouched. These refinements are grounded in fuzzy logic principles, where the square root can amplify a value and a higher-order root can modulate its impact.

IV. **Design of a Comprehensive Fuzzy Inference System (FIS):**

   We developed a fuzzy inference system to process all three sets of scores belonging to different mentioned refinements. The constructed FIS comprises comprehensive rules to produce a final, continuous sentiment score for each review, offering an expert-like evaluation that captures both overall sentiment and its intensity.

V. **Comparative Analysis Against Customer Star Ratings:**

   We compared the effectiveness of the three approaches (original scores, square root refinement, and fourth root refinement) by benchmarking the FIS-derived sentiment scores against the star ratings provided by customers. Although not a perfect measure, these ratings serve as a useful proxy for satisfaction, enabling us to discuss the strengths and limitations of each method.

VI. **Investigation of Missing Star Data:**

   Our study also examines records with missing star ratings, analyzing their potential distribution based on the three different sentiment analysis approaches while also considering human insights.

VII. **Examination of the Relationship Between Review Length and Sentiment:**

   Lastly, we explored whether the number of words in a review correlates with the sentiment expressed.

## 2. Materials and methods

### 2.1. Dataset

The dataset was constructed by gathering 1266 Persian-language comments from various Iranian restaurants. To ensure comprehensiveness in sentiment representation, comments were reviewed under human supervision. Alongside the text reviews, additional data were collected for each entry, including the number of unique items purchased and star ratings provided by customers, categorized into 2, 3, 4, 5, and not recorded (missing) values. These stars can be compared with the results of sentiment analysis after conducting the study, even though it is not a perfect measure.

Given that the original comments were in Persian, the authors employed an effective AI-based tool (GPT-4O) to convert the text into English. This translation process was carefully monitored by human experts to prevent any bias or errors. We contend that this approach does not compromise the scientific rigor of the study; rather, it demonstrates that AI-driven translation, when complemented by human oversight, can effectively handle large volumes of data translation with both speed and precision.

### 2.2. Preprocessing the English text

As mentioned, in this study, Persian comments were carefully translated into English to leverage a variety of English-based analytical tools. To ensure the quality and consistency of the data for subsequent analysis, a systematic preprocessing procedure was implemented using the Pandas library in Python.

The preprocessing pipeline consists of several key steps. First, all input text is converted to lowercase to ensure uniformity across the dataset. Next, extraneous elements such as text enclosed in brackets, URLs, HTML tags, punctuation, newline characters, and any words containing numbers are removed. Following this cleaning phase, common stop-words are filtered out, thereby eliminating semantically insignificant words. Finally, the remaining words undergo stemming to reduce them to their root forms. The fully preprocessed text is then stored in a new column within the dataset, making it suitable for further analysis.

### 2.3. Sentiment Intensity Analyzer

Sentiment Intensity Analyzer from NLTK's VADER package, a rule-based tool that utilizes a pre-defined sentiment lexicon and heuristic rules, was employed to evaluate the sentiment of each review. For every text, the analyzer computes three normalized scores (positive, negative, and neutral) that together sum to one, offering a standardized snapshot of the sentiment expressed.

## 2.4. Refined Sentiment Scores based on Fuzzy Logic

Despite ML tools, VADER is a non-trainable, rule-based approach. While it performs adequately in many general contexts, its effectiveness may diminish when applied to specific industries and tailored applications such as food-related reviews. In our study, we observed that the initial sentiment scores frequently exhibited a bias toward neutrality. In numerous cases, even when a review warranted a positive or negative classification, the neutrality score was disproportionately high. This tendency undermined the reliability of the baseline scores.

To address this limitation, these initial scores serve only as a starting point before further refinement using fuzzy logic principles. The ultimate goal of our research is to employ a fuzzy inference system (FIS) to derive a single numerical value that accurately reflects the true sentiment of each review. The FIS takes as inputs the sets of sentiment scores, specifically, the positive, negative, and neutral values. Given the inadequacies and inefficiency of the raw scores produced by VADER, we developed three distinct approaches to generate three sets of sentiment scores.

In our first approach, the original sentiment scores are not changed, leaving the score produced by the Sentiment Intensity Analyzer intact for the following FIS. In the second approach, while the neutral scores remain unaltered, the positive and negative scores are amplified by taking their square roots. This adjustment aims to amplify the impact of positive and negative sentiments that may otherwise be overshadowed by neutral bias. Using human supervision, this approach seems rational as, based on observations, we can easily confirm the situation for those data points whose positive and negative scores were marginalized, are now magnified satisfyingly.

In the third approach, similar to the second, the neutral scores are maintained; however, the positive and negative scores are instead amplified by taking their fourth roots. Although the use of the fourth root is unconventional, it was deliberately included to assess whether further modulation of the scores might yield more accurate sentiment representation and align more with reality?

Notably, due to the nature of these transformations, the summation of the modified sentiment scores does not necessarily equal one. Nevertheless, this discrepancy does not pose an issue, as the FIS is designed to operate with any input values within the [0, 1] interval and ultimately consolidate them into a single sentiment score for each review.

Remarkably, although approaches 2 and 3 successfully amplified the positive and negative scores for a significant portion of the data, there were instances where even these modifications proved insufficient. Examples of such cases will be detailed in the discussion section.

For better clarity, Figure 1 provides a comparative analysis and depiction of the three approaches, showcasing their respective advantages. The figure clearly illustrates instances where sentiments that should have been distinctly positive or negative were erroneously skewed towards neutrality, and how approaches 2 and 3 effectively amplified the correct scores. However, it must once more be emphasized that these scores are all initial perspectives, and the final sentiment score will be computed by the FIS from these scores.

**EXAMPLE 1**

Persian: بدترین سوخاری ... واقعا بدرد نخور ... به جای فیله فقط پودر سوخاری بود

English translation: The worst fried chicken... truly useless... instead of fillet, it was just breading powder.

| | Positive | Neutral | Negative |
|---|---|---|---|
| Approach 1 | 0.000 | 0.504 | 0.496 |
| Approach 2 | 0.000 | 0.504 | 0.704 |
| Approach 3 | 0.000 | 0.504 | 0.839 |

**EXAMPLE 2**

Persian: من از دیشب تا الان ۳ بار زینگر سفارش دادم انقدر خوب بود، ولی الان نون سیر هم گرفتم، به حدی افتضاح بود که نتونستم بخورم، مگه قورمه سبزیه؟ چرا سبزی سرخ کرده گذاشتین تو نون سیر؟ نون سیرتون به شدت افتضاح بود، من تاحالا نشده بود غذایی که میگیرم رو نخورم، همیشه خوردم، ولی نون سیر اصلا قابل خوردن نبود انقدر که افتضاح بود

English translation: Since last night, I've ordered Zinger three times because it was so good. But now I also got the garlic bread, and it was so terrible that I couldn't eat it. Is it supposed to be like Ghormeh Sabzi? Why did you put fried herbs in the garlic bread? Your garlic bread was absolutely awful. I've never had a meal I couldn't eat before; I always finish my food. But the garlic bread was so bad it was completely inedible.

| | Positive | Neutral | Negative |
|---|---|---|---|
| Approach 1 | 0.113 | 0.814 | 0.073 |
| Approach 2 | 0.336 | 0.814 | 0.270 |
| Approach 3 | 0.580 | 0.814 | 0.520 |

**EXAMPLE 3**

Persian: تنها فست فودی که حتی اگه بخوام هم نمیتونم ازش ایراد بگیرم، یعنی هرچی فکر کنم نمیتونم یدونه ایراد پیدا کنم واقعا ممنون که انقدر به مشتری احترام میذارین

English translation: The only fast-food place where I can't find any fault, even if I try. No matter how much I think, I can't find a single issue. Truly, thank you for showing so much respect to your customers.

| | Positive | Neutral | Negative |
|---|---|---|---|
| Approach 1 | 0.254 | 0.644 | 0.102 |
| Approach 2 | 0.504 | 0.644 | 0.319 |
| Approach 3 | 0.710 | 0.644 | 0.565 |

**EXAMPLE 4**

Persian: متاسفانه تعداد مرغ ها یک تکه کم بود، و با وجود برقراری تماس و گفتن مسئله، ایراد برطرف نشد.

English translation: Unfortunately, there was one piece of chicken missing, and despite contacting them and mentioning the issue, it was not resolved.

| | Positive | Neutral | Negative |
|---|---|---|---|
| Approach 1 | 0.000 | 0.849 | 0.151 |
| Approach 2 | 0.000 | 0.849 | 0.388 |
| Approach 3 | 0.000 | 0.849 | 0.623 |

*Figure 1 – Comparative illustration of the three sentiment-scoring approaches*

## 2.5. Fuzzy Inference System (FIS)

### 2.5.1. FIS inputs and outputs

In this section, we construct a fuzzy inference system (FIS) designed to emulate human expert decision-making through a rule-based framework. The FIS is applied uniformly across three distinct sets of sentiment scores derived from our separate refinement approaches. For each data point, we obtain three sets of scores (each comprising positive, negative, and neutral components) that are FIS inputs and are individually processed through the FIS. This system then produces a single, aggregated sentiment score for each set as an output.

### 2.5.2. Defining the membership function for inputs

In this step, we specify how each input variable (positivity score, negativity score, and neutrality score) is represented linguistically as *Low*, *Medium*, or *High*. To achieve this, we define three membership functions (two trapezoidal and one triangular) over the domain of discourse ranging from 0 to 1. As shown in Figure 2, the trapezoidal functions capture values at the extremes, whereas the triangular function centers around mid-range values.

By structuring the membership functions in this way, we can better differentiate subtle variations in each of sentiment scores. The trapezoidal functions allow for gradual transitions at both ends, accommodating slight uncertainties near the boundaries of "Low" and "High." In contrast, the triangular function provides a more focused peak for "Medium," reflecting moderate levels of sentiment intensity. These well-defined linguistic terms ensure that the fuzzy inference system can effectively capture the nuances of refined sentiment scores and subsequently generate more accurate overall sentiment assessments.

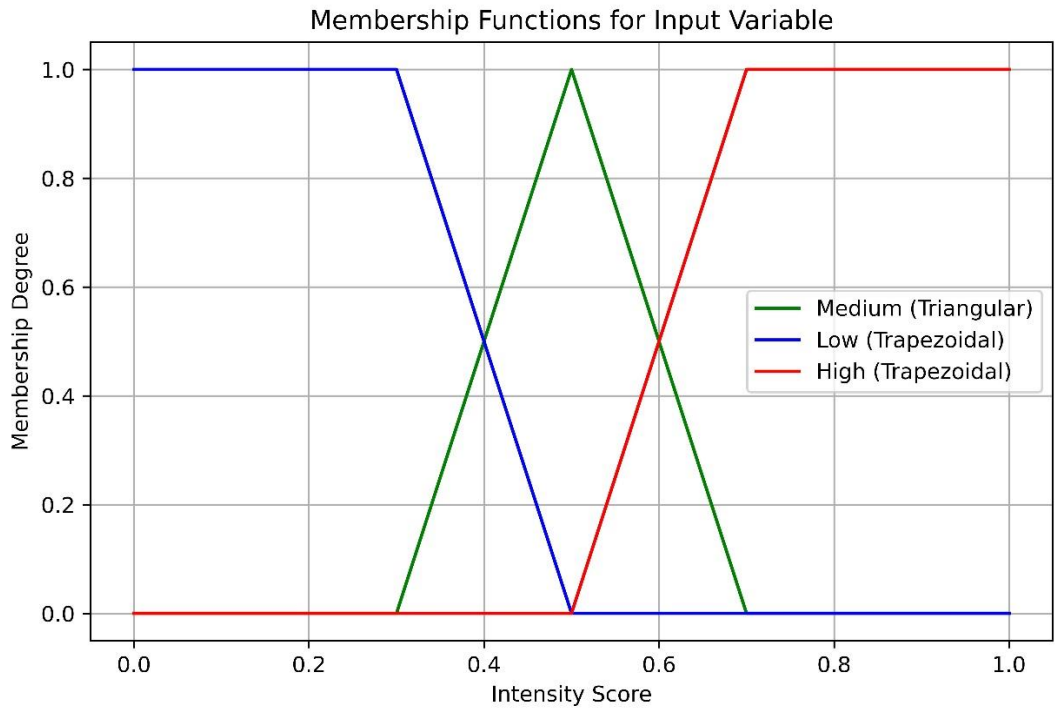

*Figure 2- membership functions for input variables (positive, negative, and neutral)*

### 2.5.3. Defining the membership function for the output

In this step, the fuzzy inference system's final sentiment evaluation (after all processes) is mapped to one of three categories (Negative, Neutral, or Positive) by assigning appropriate membership functions across the interval [0, 1]. Trapezoidal shapes are used for Negative and Positive, accommodating gradual boundaries near 0 and 1, respectively, while a triangular shape is reserved for Neutral, peaking around the midpoint.

This design clearly distinguishes the three sentiment outcomes while remaining flexible enough to handle borderline values. By defining these functions, the system can produce a single, interpretable score that concisely represents the overall sentiment of each review. For clarity, Figure 3 illustrates the output membership functions and their boundaries.

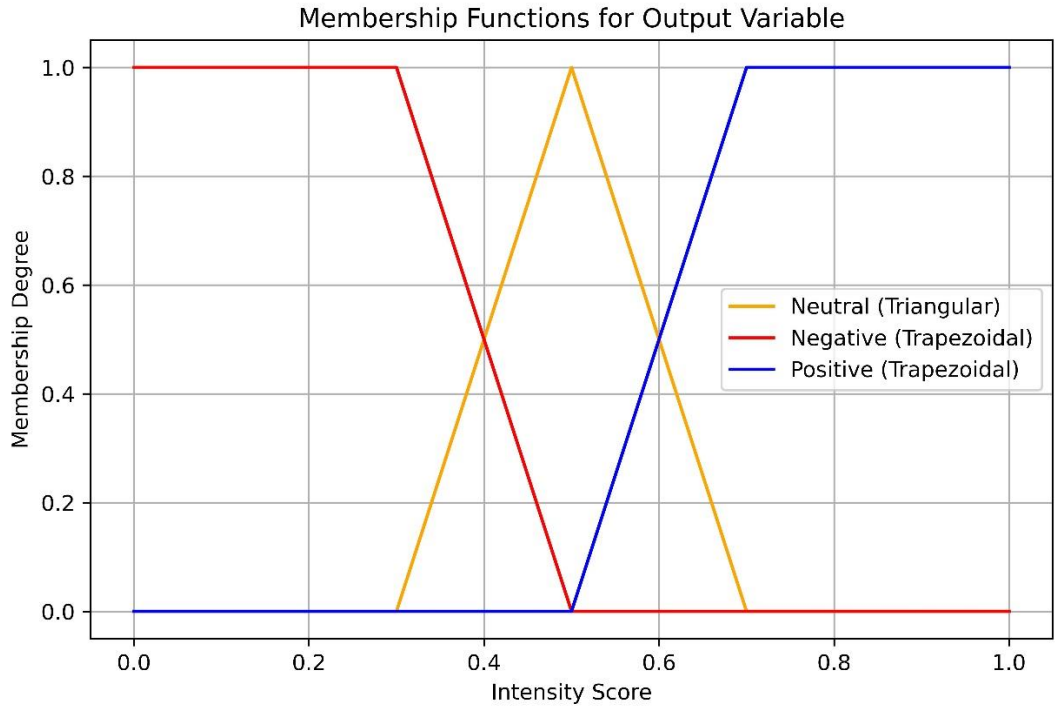

Figure 3 - membership function for the output

#### 2.5.4. Fuzzy Knowledge Base (Fuzzy Rule Base)

In this section, we establish the fuzzy knowledge base, also known as the fuzzy rule system, consisting of 27 fuzzy rules that articulate all the possible relationships between the input variables to generate the output. Each rule, expressed in the form "IF … THEN …", is defined using the "Skfuzzy" library of Python programming language.

All 27 rules are shown in Table 1, and the connector between conditions is "AND". For instance, rule 1 specifies that if the positive, negative, and neutral sentiment inputs are all classified as "low," then the overall sentiment should be interpreted as "neutral." These rules employ the minimum operator to combine multiple antecedents (T-norm) and the maximum operator for aggregating the outputs (S-norm), thereby ensuring a robust inference process.

Designing an effective rule-based (knowledge-based) system necessitates satisfying three key conditions: completeness, consistency, and continuity (Cherkassky, 1998). Completeness is achieved when every possible scenario is addressed by the rule base, ensuring that no input condition lacks a corresponding rule. In our system, this criterion is met by considering all possible combinations of input sentiment levels, namely, low, medium, and high, which yield $3^3$ (27) distinct rules. Consistency, on the other hand, is maintained when no two rules with identical antecedents produce conflicting outcomes, which did not take place. Continuity is ensured when small variations in the input lead to correspondingly gradual changes in the output, avoiding abrupt transitions in the inferred sentiment.

When constructing a rule-based system, it is crucial to consider the nature of the underlying problem. As discussed in Section 2.3 and illustrated in Figure 1, many data points—regardless of the chosen approach—are disproportionately skewed toward neutrality. Consequently, in addition to refining the sentiment scores in Section 2.4, we introduced rules within the FIS that place greater emphasis on positive and negative sentiments. This strategy should help mitigate the bias introduced by VADER's limitations, as demonstrated by rules such as 6, 12, and 18.

By systematically capturing all combinations of input levels, our fuzzy knowledge base effectively synthesizes a single, precise sentiment score from the refined sentiment inputs. As evidenced in Table 1, the logical progression among rules confirms that the system does not exhibit any paradoxical or discontinuous behavior. This structured approach guarantees that the inference system operates in a complete, consistent, and continuous manner, thereby enhancing the reliability and accuracy of the overall sentiment evaluation.

Table 1 – The comprehensive sets of rules for designing the rule-based (knowledge-based) system

| Rule | Inputs / Antecedents | | | Results |
|---|---|---|---|---|
| | Positive | Negative | Neutral | |
| 1 | low | low | low | neutral |
| 2 | low | low | medium | neutral |
| 3 | low | low | high | neutral |
| 4 | low | medium | low | negative |
| 5 | low | medium | medium | negative |
| 6 | low | medium | high | negative |
| 7 | low | high | low | negative |
| 8 | low | high | medium | negative |
| 9 | low | high | high | negative |
| 10 | medium | low | low | positive |
| 11 | medium | low | medium | positive |
| 12 | medium | low | high | positive |
| 13 | medium | medium | low | neutral |
| 14 | medium | medium | medium | neutral |
| 15 | medium | medium | high | neutral |
| 16 | medium | high | low | negative |
| 17 | medium | high | medium | negative |
| 18 | medium | high | high | negative |
| 19 | high | low | low | positive |
| 20 | high | low | medium | positive |
| 21 | high | low | high | positive |
| 22 | high | medium | low | positive |
| 23 | high | medium | medium | positive |
| 24 | high | medium | high | positive |
| 25 | high | high | low | neutral |
| 26 | high | high | medium | neutral |
| 27 | high | high | high | neutral |

### 2.5.5. Building the FIS and Data Preprocessing

The FIS is constructed in this phase using the defined fuzzy rules and membership functions. Leveraging the FIS system similar to the Mamdani-Min FIS, each If-then rule is processed by combining the antecedents using the minimum operator and aggregating the outputs using the maximum operator. The system is implemented with the fuzzy library, where a control system is built from the rule base and then simulated to process input data. For each record, a set of sentiment scores (positive, negative, and neutral) from a particular approach (1,2, or 3) is input to the FIS, and the FIS computes a single overall sentiment score via centroid defuzzification.

### 2.6. Comparison and Analysis

After processing each set of refined sentiment scores with FIS, a final overall sentiment score is computed for each approach. Consequently, three distinct overall scores are obtained corresponding to the original, square-root refined, and fourth-root refined methods. As depicted in Figure 3, the fuzzy output is defined over a range of [0, 1] with the membership functions set as follows: negative sentiment covers 0–0.5, neutral sentiment spans 0.3–0.7, and positive sentiment extends from 0.5–1. However, after achieving the FIS score, only for evaluation and comparison purposes, we designate narrower dominant ranges: scores between 0 and 0.4 indicate negative sentiment, scores from 0.4 to 0.6 indicate neutral sentiment, and scores between 0.6 and 1 indicate positive sentiment. We can state that, at this point, although fuzzy logic is employed to handle the inherent uncertainty and ambiguity in the data, for evaluation and comparison, we focus on the dominant score as a classification tool to assess the effectiveness of the employed methods.

To further evaluate these approaches, the dominant sentiment scores are compared with the star ratings provided by users, despite the inherent subjectivity of such ratings. For each refinement approach, three pie charts, referring to three dominant sentiments, are presented to illustrate the distribution of star ratings within each dominant sentiment category. Additionally, data points with missing star ratings are grouped based on their computed sentiment and subsequently reviewed and compared through human supervision. This methodology can facilitate a robust comparison of the three approaches and demonstrate its potential applicability to larger datasets, such as those derived from social media, by allowing for the sampling of random batches for human evaluation.

## 3. Results and analysis

For each approach, the corresponding sentiment scores for each approach are processed by the FIS, resulting in a final overall sentiment value. While it is not feasible to display the outcomes for all 1,266 records, Table 2 presents the results for 30 selected entries that convey both the effectiveness of employed approach as well as minor occasions where it underperformed. This table lists the final FIS-generated scores for each of the three approaches, alongside the star ratings assigned to each record. Additionally, the restaurant names have been anonymized to safeguard privacy, even though these comments are publicly available on Iranian websites.

*Table 2.* 30 reviews with their FIS overall scores and user ratings across all three approaches.

| Row | Review / Comment | Approach 1 FIS score | Approach 2 FIS score | Approach 3 FIS score | User's Stars (Rank) |
|---|---|---|---|---|---|
| 1 | The worst fried chicken... truly useless... instead of fillet, it was just breading powder. | 0.209549 | 0.204982 | 0.204982 | NAN |
| 2 | The burgers were delicious and had a very nice charcoal flavor. Unfortunately, the chicken had an odor. You should reconsider the quality of your fries as well; they are not crispy or fresh at all. | 0.668986 | 0.774994 | 0.774994 | 2 |
| 3 | The pizza was burnt, had no cheese, and was dry. It was the worst pizza I've ever had. | 0.368326 | 0.223633 | 0.218896 | 3 |
| 4 | The meat of the fried wings had a stale smell; we threw it away. | 0.500000 | 0.500000 | 0.500000 | 2 |
| 5 | The bread was stale, the fillet quality was very low, and the fried chicken breast was extremely old. | 0.500000 | 0.290037 | 0.212540 | NAN |
| 6 | The menu said thigh and breast, but they only brought the thigh. I wish I hadn't ordered. | 0.500000 | 0.795833 | 0.795833 | NAN |
| 7 | Thank you for your attention. It was exactly as requested, and it arrived much earlier than the estimated time. My child had already finished their meal while the website was still showing the preparation time. It arrived exactly 25 minutes earlier than stated, and it was very fresh and high quality. Thanks again for the quick delivery and serving such warm and delicious food. | 0.500000 | 0.782536 | 0.795833 | 5 |
| 8 | First of all, it was only thighs, and one of the thighs was almost completely undercooked, only browned on the surface. | 0.500000 | 0.500000 | 0.500000 | 3 |
| 9 | Since last night, I've ordered Zinger three times because it was so good. But now I also got the garlic bread, and it was so terrible that I couldn't eat it. Is it supposed to be like Ghormeh Sabzi? Why did you put fried herbs in the garlic bread? Your garlic bread was absolutely awful. I've never had a meal I couldn't eat before; I always finish my food. But the garlic bread was so bad it was completely inedible. | 0.500000 | 0.586514 | 0.599125 | 3 |
| 10 | The Caesar salad was just lettuce! | 0.500000 | 0.500000 | 0.500000 | 3 |
| 11 | Average. | 0.500000 | 0.500000 | 0.500000 | 4 |
| 12 | I didn't expect this at all; it was terrible. The chicken smelled like chicken! | 0.500000 | 0.787460 | 0.795833 | NAN |
| 13 | The fillets were very thin and small, mostly breading. Also, if exactly 12 pieces are to be served, they should be proper portions, not finger food-sized fillets. I deeply regret my order and will not buy again. | 0.500000 | 0.352814 | 0.223633 | NAN |
| 14 | The vegetable pizza was truly tasteless and flavorless. | 0.500000 | 0.500000 | 0.500000 | 3 |
| 15 | The description for the 3-piece fried chicken states "thigh and breast without wings"! But you only included 2 upper thighs and one drumstick! No breast!!! Also, the pack didn't include a knife and fork. | 0.500000 | 0.500000 | 0.500000 | 2 |
| 16 | You give a 20% discount but don't cut the quality by 50%. | 0.406729 | 0.223633 | 0.213403 | NAN |
| 17 | The stromboli box was open, and the cheese on the top of the burger and the top layer of the bread had come off and mixed with the rest of the box contents. One of the sauces was split into three parts. I never expected such poor packaging and delivery quality from you, and I have never experienced this before. | 0.500000 | 0.733333 | 0.789598 | 2 |
| 18 | True customer respect. | 0.795833 | 0.795833 | 0.795833 | 5 |
| 19 | Amazing | 0.500000 | 0.500000 | 0.500000 | 5 |
| 20 | The food was great, but the packaging was bad. | 0.579893 | 0.568687 | 0.500000 | 4 |
| 21 | It was good, but it arrived cold. | 0.775910 | 0.775910 | 0.775910 | 4 |
| 22 | The lamb kebab was tasteless and dry, and the Koobideh kebab tasted bitter from baking soda. I don't think it's as high-quality as it used to be. | 0.500000 | 0.313551 | 0.216899 | 2 |
| 23 | The chicken quality was good, but why was there an extra charge for sour orange when the dish didn't even include it? | 0.574117 | 0.757302 | 0.788960 | 3 |
| 24 | Excellent, excellent. | 0.795833 | 0.795833 | 0.795833 | 5 |
| 25 | The Taftoon bread under the kebab got soggy and turned into a bread stew. | 0.500000 | 0.500000 | 0.500000 | 5 |
| 26 | The food was cold due to the delay. | 0.285377 | 0.218228 | 0.218228 | 3 |
| 27 | It didn't seem fresh. | 0.788533 | 0.788533 | 0.788533 | 3 |
| 28 | *** doesn't need an introduction—it's one of the best. Unfortunately, the owner of this brand recently passed away. May their soul rest in peace, and I hope this restaurant continues its great work with the same quality. | 0.634539 | 0.776367 | 0.780658 | 5 |
| 29 | I'm giving a full score because the kebab was good and tasty, but the soup wasn't—it only tasted like flour. | 0.604617 | 0.776367 | 0.785076 | 5 |
| 30 | The food was completely cold. I'm not sure if it was the delivery or the restaurant's fault. | 0.500000 | 0.438711 | 0.449602 | 4 |

Table 2 reveals several noteworthy insights. In the first place, many samples demonstrate the effectiveness of our methodology. For example, rows 3 and 5 show that the refined approaches successfully amplified the inherent negativity of the reviews. In row 7, while approach 1 erroneously maintained a neutral score, approaches 2 and 3 shifted the sentiment toward positivity, as intended.

Similarly, in row 13—where the review explicitly conveys negativity—approach 1 yielded a neutral score, whereas approaches 2 and 3 effectively adjusted the scores to reflect the true negative sentiment. Rows 16 and 28 further illustrate this point: the refined methods accurately captured and intensified the inherent negative sentiment in row 16 and the positive sentiment in row 28, in contrast to the less representative output from approach 1. Moreover, in row 30, where a review notes cold food and slight customer dissatisfaction, approach 1 returned a neutral score, while approaches 2 and 3 successfully shifted the score to indicate slight negativity. These examples collectively highlight the practical benefits and enhanced sensitivity of the refined approaches.

Nevertheless, some limitations were observed that warrant consideration. In some cases, none of the employed approaches succeeded in altering an erroneously detected neutrality. For example, in rows 4, 8, and 14, all approaches consistently yielded a score of 0.5, despite expectations of negative sentiment. This shows some inefficiencies that are discussed further in the Discussion section.

Additionally, in a few instances—such as rows 12 and 17—the refinements appeared to compromise the results. Although the true sentiment was negative, approach 1 detected neutrality, and the refined methods further shifted the sentiment toward positivity, likely due to over-amplification of the opposite sentiment. However, these occurrences are relatively infrequent compared to the overall improvements achieved by the refined methods. Therefore, we assert that the benefits of the employed techniques significantly outweigh their limitations. Additionally, potential solutions to mitigate erroneous over-amplification are further discussed in the discussion and future work section, ensuring that the current methodology remains advantageous.

Figure 4 depicts the distribution and frequency of data points across different sentiment groups for all three approaches. In this figure, scores from 0 to 0.4 represent dominant negativity, 0.4 to 0.6 denote a predominance of neutrality, and 0.6 to 1 indicate prevalent positivity. The figure further illustrates the proportion of star ratings associated with each sentiment group for every approach, even though this is not a perfect assessment.

Figure 4 illustrates that Approach 1—where no modifications are applied—faces challenges in correctly classifying a substantial number of data points as either positive or negative. Specifically, 779 out of 1266 data points are categorized as neutral, contradicting human judgment of the data. Despite this tendency to over-assign reviews to the neutral category, Approach 1 demonstrates relatively greater accuracy among data classified as positive and negative sentiments.

In Approach 1, low star ratings (2 or 3) contribute less to the positive category than in the other two approaches, while high star ratings (4 or 5) account for a significantly larger proportion of the positive group. Additionally, within the negative sentiment category, Approach 1 shows a lower contribution of high star ratings (4 or 5). These findings indicate that although Approach 1 struggles to reallocate neutral data points into positive or negative categories, its classifications within these two groups are more precise. Furthermore, while the total number of neutral classifications varies among the three methods, the distribution of star ratings within the neutral group remains relatively consistent across all approaches, rendering this factor insignificant.

Another notable observation from Figure 4 is that, although Approaches 2 and 3 both refine and amplify positive and negative sentiments using square-root and fourth-root transformations, respectively, their performance appears to be nearly identical. That is, the distribution and frequency of data across the defined sentiment categories exhibit no significant differences between the two approaches. This indicates that, within the scope of this study, using the fourth root does not offer a clear advantage over the square root. Therefore, adopting the square root, which aligns more closely with conventional fuzzy logic principles, appears to be the more practical choice.

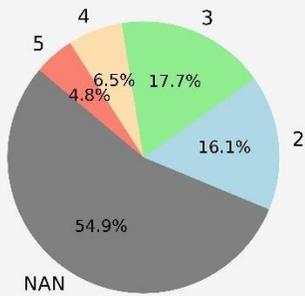
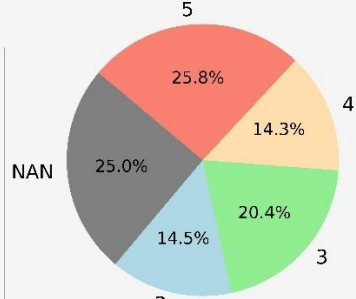
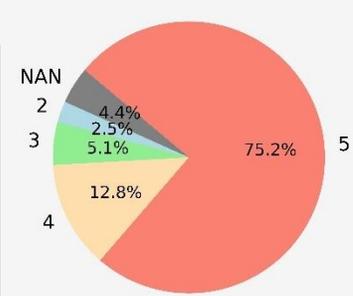
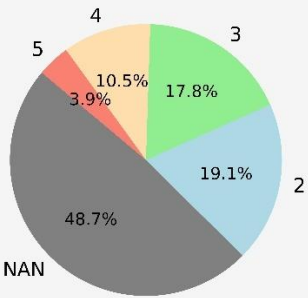
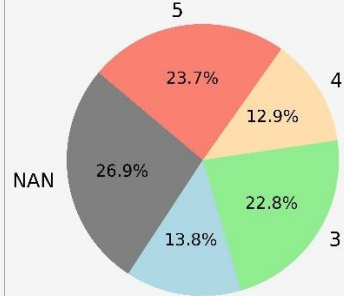
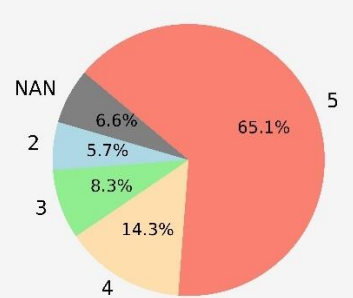
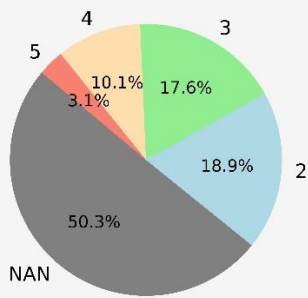
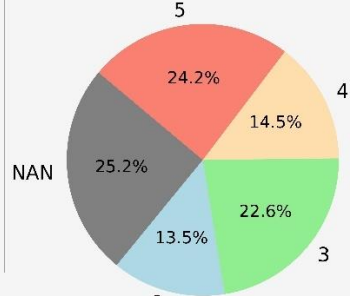
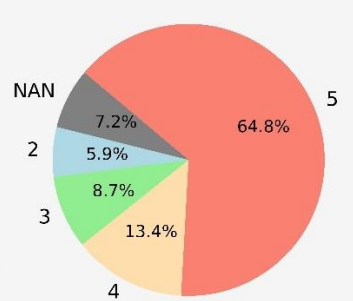

*Figure 4.* Comparison of sentiment group distributions and star ratings across approaches.

The distribution of missing data—cases where no star rating was provided by the users—was also examined. In the first place, a preliminary human assessment suggests that most of these entries exhibit either negative or positive sentiments, with negative being the most prevalent, followed by positive, and then neutral. In regard with the conducted study, Figure 5 illustrates the predicted sentiment distribution for data points with missing star ratings across the three approaches. As depicted in Figure 5, although none of the three approaches fully minimizes the neutral category, Approaches 2 and 3 outperform Approach 1 by assigning more entries to negative and positive sentiments, aligning more with reality. Moreover, the figure reinforces that Approaches 2 and 3 yield comparable results, corroborating the earlier observation that there is no substantial difference between them.

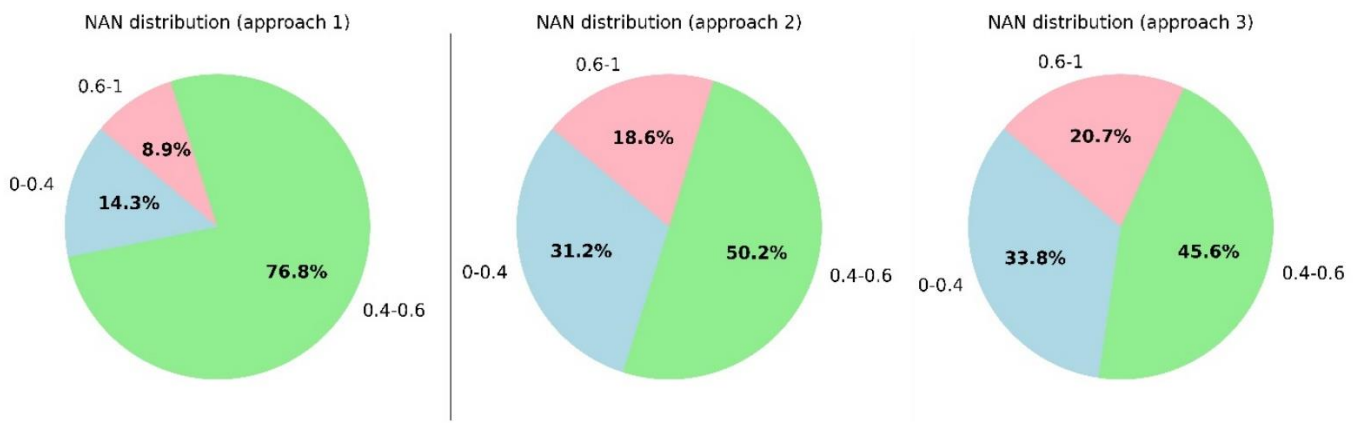

**Figure 5.** Predicted sentiment distribution for data with missing star ratings across the three approaches

We also conducted an in-depth study using a confusion matrix and heat map to examine the differences between Approaches 2 and 3 relative to Approach 1. Figure 6 illustrates how Approaches 2 and 3 adjust the sentiment labels initially assigned by Approach 1: the left panel compares Approach 1 with Approach 2, while the right panel contrasts Approach 1 with Approach 3.

In Approach 1, a considerable number of reviews are erroneously classified as neutral. By contrast, Approaches 2 and 3 reassign many of these reviews to more appropriate positive or negative categories. The red and yellow crosses on the charts highlight undesirable reclassifications, such as shifting positive or negative to neutral (yellow crosses) or even worse, to the opposite category (red crosses). Their limited or absent presence confirms that the refined methods largely avoid severe errors. Additionally, the blue arrows indicate that a significant portion of data transitions from neutrality to either positive or negative, underscoring the benefits of these refinements.

To conclude Figure 6, both Approaches 2 and 3 demonstrate minimal misclassification of positive or negative sentiments into other groups while effectively reallocating a substantial portion of neutral data into the correct categories. This consistency supports the conclusion that amplifying sentiment scores, whether via a square-root or fourth-root transformation, is a viable solution, although approaches 2 and 3 are somehow similar.

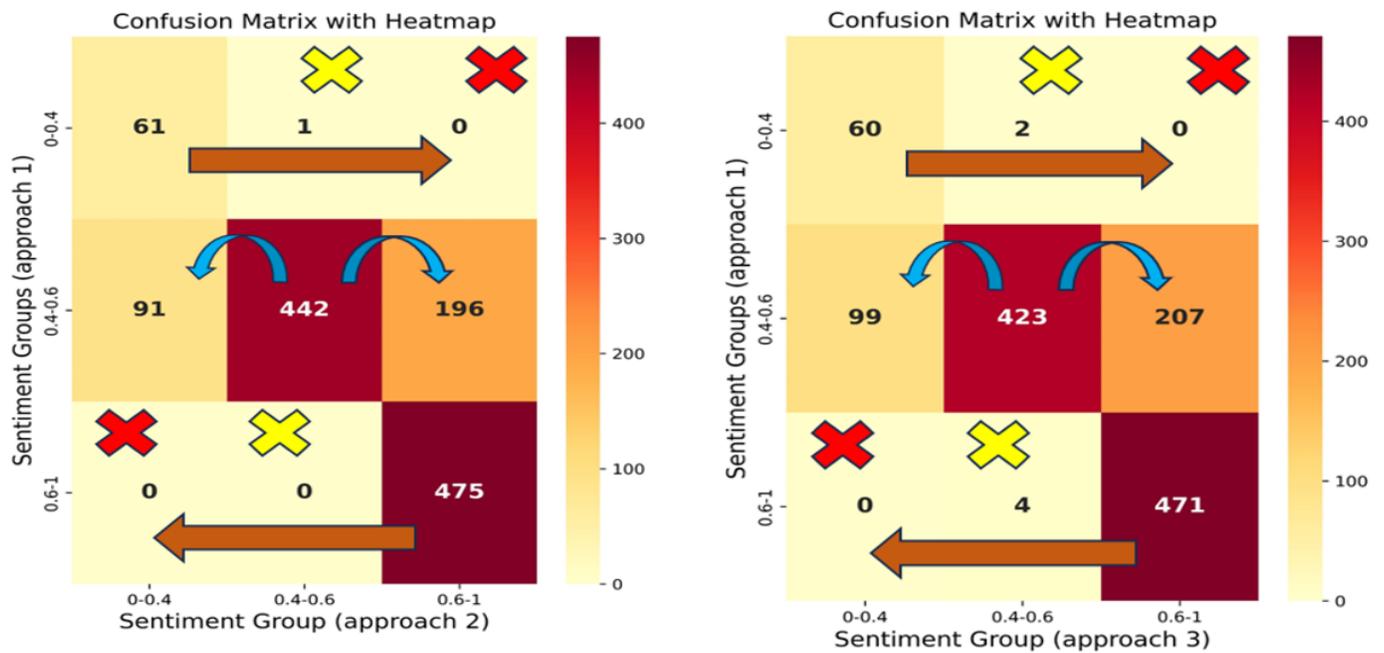

*Figure 6.* Confusion matrices showing reclassification from Approach 1 to 2/3.

A simple study was also conducted to examine whether there is a significant correlation between the number of words in a user's comment and the star rating they assigned. If such a correlation were found, we aimed to investigate whether similar patterns emerged across the three approaches. However, the analysis revealed no substantial correlation, suggesting that, at least in the current study, the length of a comment does not directly influence its sentiment. Figure 7 illustrates these findings.

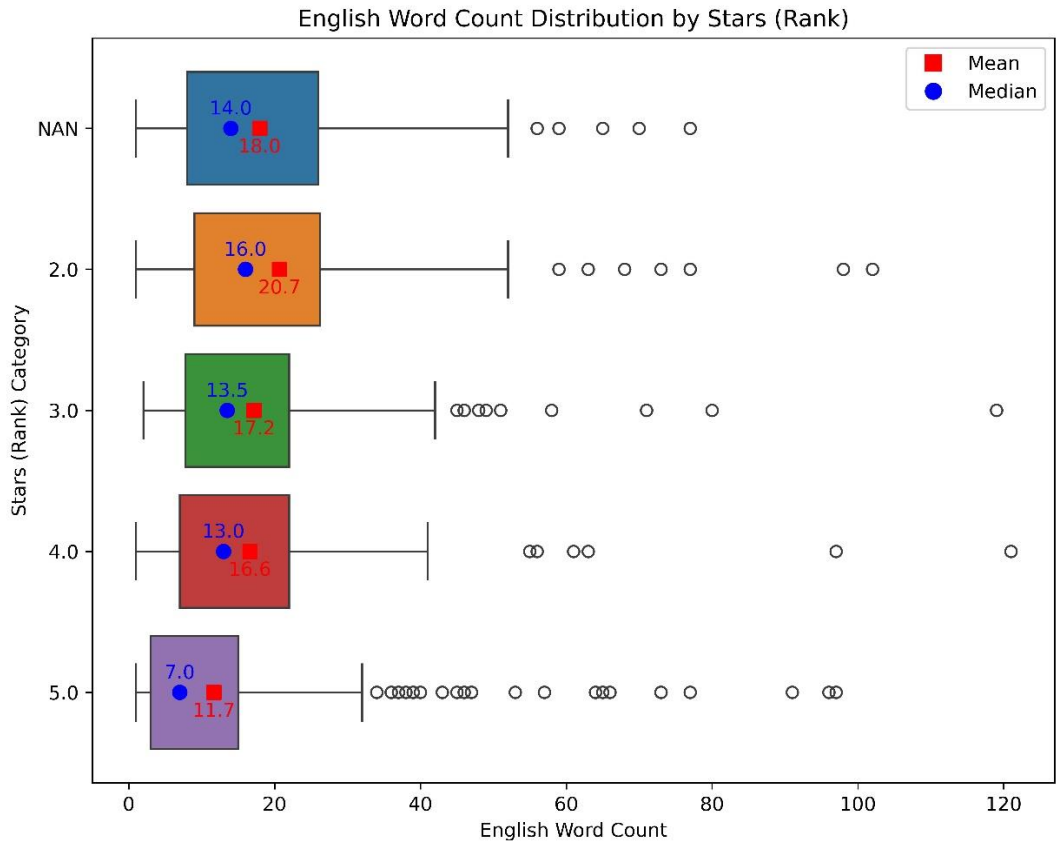

*Figure 7. Relationship between English word count and star ratings*

## 4. Discussion and Future Works

### 4.1. Effectiveness of Fuzzy Logic in Sentiment Analysis

According to results, the application of fuzzy logic in sentiment analysis has proven highly effective in addressing the inherent uncertainties present in textual reviews. Conventional sentiment analysis models, such as VADER, often struggle with reviews that contain mixed or nuanced sentiments. By leveraging FIS and a refined fuzzy-based perspective, this study was able to produce sentiment scores that better reflected the complexity of user opinions. Unlike rigid classification methods, which force text into distinct categories, the fuzzy approach allowed for a more continuous representation of sentiment, making the analysis more aligned with human perception.

### 4.2. comparison of Sentiment Refinement Approaches

Three different approaches were used in this study: the original VADER-generated scores, the square root refined scores, and the fourth-root refined scores. The original approach exhibited a strong tendency to classify a large portion of reviews as neutral, failing to capture the true sentiment intensity. The square-root and fourth-root transformations were applied to mitigate this effect by amplifying positive and negative scores, thereby reducing the neutral bias. The results indicate that while both transformations significantly improved sentiment differentiation, the square root method provided a more balanced refinement, avoiding potential over-amplification observed in some cases using the fourth root. Overall, these refinements enhanced sentiment classification accuracy and better aligned with customer star ratings.

### 4.3. Challenges and Limitations

Despite the success of the refined approaches, some challenges remain. The persistence of neutrality bias in certain cases suggests that even fuzzy logic cannot entirely resolve sentiment ambiguity. Some reviews remained classified as entirely neutral despite their evident positive or negative intent. Additionally, in a few instances, sentiment over-amplification led to incorrect classifications. This was particularly noticeable when a review contained both positive and negative elements, as the refinement process sometimes disproportionately enhanced the erroneous aspect over the other. Addressing these issues could require further optimization of the fuzzy membership functions, knowledge-based systems, and often, marginally customizing the sentiment word dictionary for the specific tasks.

### 4.4. Impact of Missing Star Ratings on Sentiment Evaluation

Although user reviews (given stars) do not provide the best assessment method, the key challenge in sentiment analysis is still handling reviews without explicit user ratings. The outcomes demonstrated that the refined approaches were more effective in assigning appropriate sentiment scores to such cases, although none of them was up to the mark. Unlike the original VADER method, which frequently misclassified these reviews as neutral, the refined models provided a better solution, highlighting the potential of fuzzy logic to be partially successful in intensity-based sentiment classification tasks, even in the absence of external rating benchmarks.

### 4.5. Relationship Between Review Length and Sentiment

An exploratory analysis was conducted to determine whether review length correlated with sentiment intensity. The findings suggest that no strong relationship exists between the number of words in a review and the sentiment expressed. This, at least in our study, challenges a common assumption that longer reviews provide clearer sentiment indications. The lack of correlation implies that sentiment analysis should rely primarily on textual content rather than length as a determinant of sentiment strength. Future studies can explore linguistic structures or sentiment-weighting mechanisms to refine this aspect further.

### 4.6. Potential Real-World Applications

The findings of this study offer substantial practical benefits, particularly for businesses that depend on customer feedback, such as restaurants and online platforms. By incorporating sentiment analysis enhanced with fuzzy logic refinements, businesses can attain a deeper and more precise understanding of customer sentiments. As highlighted, fuzzy logic not only identifies the overall sentiment but also captures its intensity, a critical factor in distinguishing varying degrees of customer satisfaction. This capability facilitates automated review analysis, enables more targeted service enhancements, and supports real-time decision-making for addressing customer concerns. Furthermore, e-commerce platforms and recommendation systems can leverage these refined sentiment evaluations to provide more accurate product suggestions, ultimately improving user engagement and overall satisfaction.

### 4.7 Future Research Directions

While this study has demonstrated the advantages of fuzzy logic in sentiment analysis, several avenues for future research remain open. Addressing the observed challenges could further enhance the robustness and accuracy of sentiment classification. One key limitation identified was the occasional misclassification of inherently positive or negative sentiments as pure neutral (0.5 score). Observations state that this primarily occurred when sentiment-bearing words were domain-specific rather than general sentiment words (e.g., "flavorless, undercooked, burnt" in the food industry). To mitigate this, we propose fine-tuning VADER (or deep learning models, in other studies) with a specialized domain-specific lexicon. This refinement would help reduce erroneous neutral classifications by improving sensitivity to industry-specific sentiment indicators.

Additionally, in sentiment refinement approaches (Approaches 2 and 3), we observed occasional misclassifications where sentiments were over-amplified in the wrong direction. To address this, we propose integrating fuzzy logic with a neural network explicitly designed for sentiment analysis. Although the neural network categorizes the sentiment as 0 or 1, it can work as a filter to determine which sentiments should be amplified and which one should not, ultimately offering a superior refinement.

Furthermore, expanding the dataset to include multiple industries, languages, and larger corpora would enhance the generalizability of the proposed method. Finally, integrating real-time sentiment tracking into business intelligence tools could provide valuable insights for customer-centric decision-making. By addressing these challenges and expanding the scope of our methodology, future research can further improve sentiment analysis models, making them more precise, adaptable, and applicable across various domains.

## 5. Conclusion

This study introduced an enhanced sentiment analysis framework that integrates fuzzy logic principles with conventional sentiment analysis techniques to address key challenges in sentiment classification. Unlike traditional classification methods that adopt a binary perspective, fuzzy logic enables the incorporation of sentiment intensity, allowing for a more nuanced representation of emotions. The dataset used comprised 1266 reviews from Iranian restaurants, translated into English via AI, along with their corresponding user-assigned star ratings. The initial sentiment analysis was conducted using VADER, a conventional tool that assigned positivity, negativity, and neutrality scores to each review. However, due to inefficiencies such as a noticeable bias toward neutrality, the results often failed to accurately capture the true sentiment. To refine these scores, two additional approaches were introduced, applying square-root and fourth-root transformations to amplify positive and negative scores while keeping neutrality unchanged. This process generated three distinct sentiment score sets. An FIS was then designed with a comprehensive rule base to process these scores. Each sentiment score for each approach is fuzzified into "low," "medium," or "high", representing FIS inputs, and then processed using a knowledge base similar to the Mamdani-Min inference system. The output is then defuzzified through a centroid defuzzification to generate a single continuous value between 0 and 1, reflecting both the sentiment and its intensity.

The comparative analysis of sentiment scores demonstrated that the combination of fuzzy logic refinements and FIS significantly improved sentiment classification, particularly in cases where VADER's neutrality bias had previously overshadowed sentiment intensity. While the refined methods effectively captured the underlying emotions in most cases, a few instances of over-amplification or insufficient refinement were observed, which were further discussed along with potential solutions. Additionally, an analysis of missing star ratings revealed that the refined approaches more accurately classified sentiments compared to the baseline method. A confusion matrix comparison further confirmed the effectiveness of the refinements, with minimal misclassifications and substantial improvements. Lastly, a correlation analysis between review length and sentiment revealed no meaningful relationship, suggesting that sentiment intensity is independent of text length.

Overall, the findings highlight the advantages of integrating fuzzy logic into sentiment analysis, particularly in handling sentiment complexity and intensity. While the proposed framework significantly improved sentiment classification accuracy, minor challenges such as occasional misclassification due to over-amplification remain. Future research could explore deep learning integration, optimization of fuzzy rule sets, and application across diverse industries and languages.

**CRediT authorship contribution statement**

**Shayan Rokhva:** Writing – review & editing, Writing – original draft, Visualization, Validation, Software, Resources, Project administration, Methodology, Investigation, Formal analysis, Data curation, Conceptualization. **Babak Teimourpour:** Writing – review & editing, Validation, Supervision, Project administration. **Romina Babaie**: Visualization, Validation